\documentclass[sigconf,authorversion]{acmart}
\AtBeginDocument{%
  \providecommand\BibTeX{{%
    \normalfont B\kern-0.5em{\scshape i\kern-0.25em b}\kern-0.8em\TeX}}}

\copyrightyear{2024}
\acmYear{2024}
\setcopyright{rightsretained}
\acmConference[GeoAI'24]{7th ACM SIGSPATIAL International Workshop on AI for Geographic Knowledge Discovery}{October 29-November 1, 2024}{Atlanta, GA, USA}
\acmBooktitle{7th ACM SIGSPATIAL International Workshop on AI for Geographic Knowledge Discovery (GeoAI'24), October 29-November 1, 2024, Atlanta, GA, USA}
\acmDOI{10.1145/3687123.3698286}
\acmISBN{979-8-4007-1176-3/24/10}

\usepackage{listings}
\usepackage{graphicx}
\usepackage{booktabs}
\usepackage{float}

\newfloat{lstfloat}{htbp}{lop}
\floatname{lstfloat}{Listing}

\definecolor{codegreen}{rgb}{0,0.6,0}
\definecolor{codegray}{rgb}{0.5,0.5,0.5}
\definecolor{codepurple}{rgb}{0.58,0,0.82}
\definecolor{backcolour}{rgb}{0.95,0.95,0.92}

\lstdefinestyle{mystyle}{
    backgroundcolor=\color{backcolour},   
    commentstyle=\color{codegreen},
    keywordstyle=\color{magenta},
    numberstyle=\tiny\color{codegray},
    stringstyle=\color{codepurple},
    basicstyle=\ttfamily\footnotesize,
    breakatwhitespace=false,         
    breaklines=true,                 
    captionpos=b,                    
    keepspaces=true,                 
    numbers=none,                    
    numbersep=5pt,                  
    showspaces=false,                
    showstringspaces=false,
    showtabs=false,                  
    tabsize=2
}

\begin{document}

\title{Evaluation of Code LLMs on Geospatial Code Generation}

\author{Piotr Gramacki}
\orcid{0000-0002-4587-5586}
\affiliation{%
  \institution{Wrocław University \\ of Science and Technology}
  \department{Department of Artificial Intelligence \\ Kraina.AI}
  \city{Wrocław}
  \country{Poland}
}
\email{piotr.gramacki@pwr.edu.pl}

\author{Bruno Martins} 
\orcid{0000-0002-3856-2936}
\affiliation{%
    \institution{INESC-ID \& LUMLIS {\footnotesize(Lisbon ELLIS Unit)} }  
    \institution{Instituto Superior Técnico \\ University of Lisbon}
    \city{Lisbon}
    \country{Portugal}
} 
\email{bruno.g.martins@tecnico.ulisboa.pt}

\author{Piotr Szymański}
\orcid{0000-0002-7733-3239}
\affiliation{%
  \institution{Wrocław University \\ of Science and Technology}
  \department{Department of Artificial Intelligence \\ Kraina.AI}
  \city{Wrocław}
  \country{Poland}
}
\email{piotr.szymanski@pwr.edu.pl}


\begin{abstract}
    Software development support tools have been studied for a long time, with recent approaches using Large Language Models (LLMs) for code generation. These models can generate Python code for data science and machine learning applications. LLMs are helpful for software engineers because they increase productivity in daily work. An LLM can also serve as a “mentor” for inexperienced software developers, and be a viable learning support. High-quality code generation with LLMs can also be beneficial in geospatial data science. However, this domain poses different challenges, and code generation LLMs are typically not evaluated on geospatial tasks. Here, we show how we constructed an evaluation benchmark for code generation models, based on a selection of geospatial tasks. We categorised geospatial tasks based on their complexity and required tools. Then, we created a dataset with tasks that test model capabilities in spatial reasoning, spatial data processing, and geospatial tools usage. The dataset consists of specific coding problems that were manually created for high quality. For every problem, we proposed a set of test scenarios that make it possible to automatically check the generated code for correctness. In addition, we tested a selection of existing code generation LLMs for code generation in the geospatial domain. We share our dataset and reproducible evaluation code on a public GitHub repository\footnote{\url{https://github.com/kraina-ai/geospatial-code-llms-dataset}}, arguing that this can serve as an evaluation benchmark for new LLMs in the future. Our dataset will hopefully contribute to the development new models capable of solving geospatial coding tasks with high accuracy. These models will enable the creation of coding assistants tailored for geospatial applications. 
\end{abstract}




\maketitle

\section{Introduction}

Large Language Models (LLMs) can nowadays be effectively used in tasks related to code generation, particularly when considering popular programming languages and domains, such as Python code for data science and machine learning~\cite{zan-etal-2023-large}. 
These capabilities can make programmers' daily work more accessible, for instance increasing expert productivity through assistance in writing code. They can also offer learning support to novice developers, returning code examples from problem descriptions in natural language. The benefits mentioned above can also extend to specialised problem domains such as geospatial data science, where problems often involve generating repetitive code for performing particular geospatial analyses and data visualisations.

Although using code LLMs for geospatial data science seems like a trivial application, we argue that there are several challenging aspects to consider. For instance, the understanding of the natural language problem descriptions can involve geospatial reasoning and world knowledge~\cite{Bhandari2023}. Geospatial data analyses also frequently involve using specialised software libraries (e.g., for geo-coding data into/from place names, performing geometric operations over geospatial regions, analysing raster data, producing map-based visualisations, etc.). The amount of code examples using these libraries, as available in the public repositories from which data are crawled to support the training of code LLMs, is probably also much smaller than the number of examples using generic data science and machine learning libraries (which are also covered in dedicated datasets used in previous work, like DS-1000~\cite{Lai2022}).

Noting the challenges mentioned earlier, our research sought answers to the following three main research questions:
\begin{itemize}
    \item[\textbf{RQ1:}] Are code generation LLMs capable of solving different  types of geospatial tasks?
    \item[\textbf{RQ2:}] Can code generation LLMs use spatial reasoning and world knowledge when solving geospatial tasks?
    \item[\textbf{RQ3:}] Considering a broad categorisation of geospatial tasks, what types of problems are currently more challenging for code generation LLMs?
\end{itemize}
We specifically conducted an evaluation of existing models for code generation on a selection of geospatial tasks that test knowledge about spatial reasoning, spatial data processing, and available tools. 

To do this, we started by defining a broad categorisation of geospatial data science tasks, considering dimensions such as the complexity of the task (e.g., single-step operations involving the processing of different representations for spatial data, versus complex analyses involving multiple steps) or the tools that can be used. We then compiled a dataset of specific coding problems, manually curating the tasks and test scenarios to match the proposed categorisation of problems.

Using the proposed dataset, we can automatically check the generated code's correctness. Therefore, we argue that one of our main contributions is a carefully designed (and extensible) benchmark for assessing code generation LLMs in the geospatial domain. We also report on a comparative study that covers a representative set of modern code generation LLMs, with results highlighting the capabilities and limitations of current state-of-the-art models.

The remainder of this paper is organised as follows. Section 2 describes related work, covering code LLMs and existing generalist benchmarks, as well as the use of LLMs for geospatial tasks. Section 3 describes the benchmark proposed in this work, detailing the proposed categorisation of problems and the construction of the dataset. Section 4 presents the results of experiments comparing different code LLMs using the proposed benchmark. Finally, Section 5 summarises the paper's main contributions and highlights possible directions for future work.

\section{Related Work}

In this section, we investigate previous work relevant to our research questions. We focus on Large Language Model (LLM) applications for general code generation tasks, to find out how such evaluations were performed in prior work. We focus primarily on work presenting code generation benchmarks, since these studies are more relevant to our research. Apart from that, we investigate the application of LLMs to problems related to the geospatial domain. This should give us a better understanding of the applications of LLMs in this domain, and their main limitations. We aim to address those issues in our benchmark.

\subsection{Large Language Models (LLMs) for Code}

Large Language Models (LLMs) have been successfully applied to code generation tasks, and early approaches have been summarised in a survey by \citet{zan-etal-2023-large}. Since then, multiple models for code generation have been introduced, sometimes as small as 3B parameters ~\cite{Roziere2023, Lozhkov2024, IBM-Granite, CodeGemma}. Also, some of the more recent generic models are also capable of solving code generation tasks~\cite{llama3modelcard, Jiang2023, Gemma}.

Evaluation of code generation capabilities requires dedicated datasets. HumanEval~\cite{Chen2021} was proposed from the need to consider an evaluation dataset of handcrafted samples which were unseen by the LLM during pre-training. It was later followed by several similar datasets, either for generic programming tasks~\cite{Austin2021, evalplus, Hendrycks2021} or for specific domains or libraries~\cite {Lai2022, Zan2022}. All these datasets share a couple of principles, e.g. stressing the importance of human validation of samples. They also provide test cases to evaluate the functional correctness of the generated code.

\subsection{LLMs for Geospatial Tasks}

In a recent survey, \citet{Wang2024} summarised studies on LLM and Generative AI (GAI) applications to the geospatial domain. The authors found a variety of applications handling raster, vector, and spatio-temporal data. The most common applications involve the use of text-based models, which we also cover in our work. 

Solving geospatial tasks with LLMs was approached differently in existing studies. For instance \citet{Zhang2023} proposed GeoGPT, which suggests a set of tools which extend LLM capabilities to perform spatial queries and analysis. \citet{Li2023} used the code generation capabilities of an LLM to split a task into sub-tasks and generate code snippets to solve them separately. These authors proposed a vision of autonomous-GIS, where non-programmers can solve complex geospatial tasks. 

Another area of research on the boundary between LLMs and GeoAI concerns assessing the geospatial awareness and knowledge obtained by the models. \citet{Bhandari2023} showed that recent LLMs have gathered geospatial knowledge, which can be extracted using proper prompting. \citet{Roberts2023} evaluated the spatial knowledge of GPT-4 on tasks that require both factual knowledge (e.g., country population or boundaries) and reasoning (e.g., routing and navigation). \citet{Manvi2023} assessed the basic geographic knowledge of LLMs and how to extract it, by enriching text prompts with OpenStreetMap data. Finally, \citet{Mooney2023} tested the geospatial capabilities of GPT models using a real GIS exam, reporting that both GPT-3 and GPT-4 passed the exam with the respective grades of D and B, on the US letter scale.

The need for dedicated models for the geospatial domain was also highlighted in multiple vision and outlook papers. The studies by \citet{Scheider2021} and \citet{Mai2021} focused on geospatial question answering. The authors highlighted the unique characteristics of geospatial data, and the fact that complex questions cannot be answered using direct knowledge retrieval solutions. These questions require some level of analytical creativity, but can yield valuable insights for users - for example, a doctor could ask for potential environmental factors influencing a patient's health. In a different study, \citet{Mai2023} argued that versatile models for the geospatial domain should focus on the multi-modal characteristics of geospatial data and tasks. The authors show that existing single-modality models can perform well on the geospatial domain, but the majority of tasks depend on multi-modal inputs.

To summarise, LLMs have been adopted in the geospatial domain, with many studies showing that those models have potential in GIS applications. Autonomous-GIS can leverage the capabilities of LLMs to understand text instructions and use tools or code to solve different tasks. However, there are still some unique challenges in the geospatial domain, and we argue that the evaluation of the code generation capabilities of modern LLMs is an essential step in the development of assistants helpful in geospatial tasks.

\section{Proposed Benchmark}

This section describes the design process of our geospatial benchmark. We propose a categorisation of geospatial problems which takes into consideration both the complexity of the problem and the tools or geospatial knowledge required to solve them. We also describe our dataset in terms of prompt format and test case creation. Finally, we provide detailed statistics for our benchmark dataset, along with examples of tasks for each category.

A first consideration was to select a programming language to be used in our benchmark. Two popular options in the GIS community are Python and the R language for statistical computing. Both would correspond good candidates for this benchmark, but we went with \textit{Python} primarily because it is present at larger scale in existing code corpora, which makes the available models better at handling this particular language~\cite{Kocetkov2022TheStack}.

\subsection{Categorisation of Geospatial Problems}

We organise our benchmark around a four-dimensional classification of problems: (1) the complexity of the task, (2) input type for geospatial information, (3) the required tools and knowledge to solve it, and (4) the framing of a task.

\paragraph{Task complexity} This is the main criterion by which we classify the considered test cases. It is defined as one of two possible values: \textit{single-step operation} or \textit{multi-step analysis}. The first one fits the tasks that require only a correct selection of a single geospatial operation. The second one covers complex tasks which require multi-step operations.

\paragraph{Input type} This dimension specifies different formats for geospatial data, which are used in a test case. We use primitive geospatial data types (such as points) either as just numbers or using dedicated objects. This could be further extended with polygons upon extending our dataset. We also cover collections of geospatial data using different file and data formats. In total, this dimension covers:

\begin{itemize}
    \item the GeoPandas library and GeoDataFrames usage~\cite{joris_van_den_bossche_2024_12625316};
    \item GeoJSON files;
    \item Shapefiles;
    \item the shapely library for geometries management;
\end{itemize}

\paragraph{Tools usage} In this dimension, we specify different resources to solve each task. We cover basic tools to perform geospatial queries and analysis. In this version of our dataset, we cover:

\begin{itemize}
    \item the OSMNX library for geocoding;
    \item the H3 library for spatial indexing;
    \item the shapely library for geometries management;
    \item the MovingPandas library for trajectory data;
\end{itemize}

\paragraph{Task framing} Finally, we propose this dimension to test the understanding of spatial operations by an LLM. We propose two approaches to frame the task description: \textit{operation} and \textit{semantic}. The operation framing clearly describes which spatial operations are expected to happen to obtain a result. We describe the desired operation using concrete geometry types and geospatial operations. The semantic framing poses a more general question and frames the task, in natural language, as a general goal to reach. We aim to use real-life examples there to see if the model knows which operations should be used to achieve this. An example of the difference between the two types could be "counting points that intersect polygons" (\textit{operation}) framed also as "counting bus stops in districts" (\textit{semantic}). We propose to frame tasks that are identical in term of operations required as either \textit{operation} or \textit{semantic} framing. The resulting code would be identical in both of those situations. We use this dimension to see if the change in training affects the capability of the LLM to complete a given task, which would mean that it does not understand the underlying spatial operations.

The aforementioned categories are not comprehensive but offer broad coverage of different operations specific to geospatial data and tasks. (i) \textit{GeoJSON} and \textit{Shapefile} are two of the most popular types of geospatial data storage. Loading spatial data formats and correctly interpreting the results is crucial for all spatial analysis tasks. That is why we included this category in our benchmark. (ii) The \textit{shapely} library offers a low-level API for geometrical operations. Geometries are the foundation of GIS, which is why construction and basic operations on geometries have found their way into our benchmark. (iii) \textit{GeoPandas} is a high-level library for geospatial data processing, which extends the popular \textit{Pandas} library. It is the main tool for loading spatial data in Python and processing them. It is also one of the most popular formats in which other geospatial libraries return results. (iv) \textit{OSMNX} is a library which handles access to data from the OpenStreetMap (OSM), which is the primary source of open data in the geospatial domain. It also enables geocoding via the Nominatim service. Geocoding is often tested as a geospatial capability of LLMs, and recent studies suggest that LLMs tend to be biased towards better performance on higher developed countries~\cite{Manvi2024}. That is why we argue that testing those models' capabilities to use specific geocoding tools is essential. (v) \textit{H3} is a spatial index proposed by Uber. It has become very popular because it allows for deterministic tessellation of space for analytical tasks. We include it in our benchmark as an example of a more specific library, to see if existing code models know its syntax and basic operations. Coding assistants for geospatial tasks should be able to use these libraries. If existing models fail to do so, this would be a valuable insight towards future improvements of geospatial code generation models. (vi) \textit{MovingPandas} is a library designed to process trajectory data, which is widely used in geospatial analysis. This library is very capable and offers compatibility with \textit{GeoPandas}, which makes it a complementary fit for this benchmark.

\subsection{Format for Code Prompts}

\begin{figure}
    \centering
    \includegraphics[width=\linewidth]{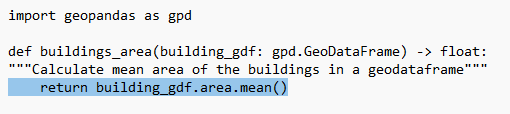}
    \caption{Example of a prompt from the dataset, with an example solution generated (the highlighted part).}
    \label{fig:example-task}
\end{figure}

Each sample in our dataset has the same prompt format, which also remains unchanged between tested models. The prompts are human-written to ensure that they were not present in any training data for existing models. We provide a function signature with typing, and a docstring with the task description. This is consistent with other studies proposing code generation datasets~\cite{Chen2021, evalplus, Austin2021, Hendrycks2021}. Some samples have an import of libraries used in typing. An example of the prompt format for our dataset is presented in Figure~\ref{fig:example-task}.

\subsection{Samples creation process}

Each task can be seen from the perspective of multiple dimensions in our categorisation - complexity of a task, instruction framing, format of the input, and tools required. We made sure that there are multiple samples for each possible class in any dimension. We start with defining a task which falls under the first dimension of our categorisation - \textit{single-step} or \textit{multi-step} operation. Having done that, we augment this task by changing the format of input parameters or framing of the task. For example we can take a task "calculate mean area of polygons" and have three variants for different input formats. Then we reformulate the instruction to use the \textit{semantic} framing and we already have $3 \times 2 = 6$ samples, after manually writing just two of them. We also swap used tools if possible. This way, we are able to obtain more hand-crafted samples in our dataset. Through our procedure, we obtained a larger dataset, at the same time having the possibility to assess performance on a single tool/data type. Another advantage of this approach is that we can calculate model performance on a single dimension, to better assess individual capabilities. An example of this an approach and its obtained samples is presented in Listing~\ref{lst:sample-augmentation}.

\begin{table}[t]
    \centering
    \caption{Summary of task categories.}
    \label{tab:sample-categories}
    \begin{tabular}{c|cc|c}
    \toprule
    Operation type  & Single    & Multi-step    & Not used \\
    \midrule
    \multicolumn{4}{c}{Input type}          \\
    \midrule
    GeoPandas       & 4         & 5             & 11 \\
    Shapefile       & 4         & 2             & 14 \\
    GeoJSON         & 4         & 5             & 11 \\
    Lat\/Lon points & 4         & 2             & 14 \\
    Shapely points  & 3         & 2             & 15 \\
    \midrule
    \multicolumn{4}{c}{Tools usage}         \\
    \midrule
    OSMNX           & 3         & 1             & 16 \\
    H3              & 2         & 1             & 17 \\
    Shapely         & 4         & -             & 16 \\
    MovingPandas    & -         & 3             & 17 \\
    \midrule
    \multicolumn{4}{c}{Task framing}        \\
    \midrule
    Operation       & 7         & 5             & 8 \\
    Semantic        & 7         & 5             & 8 \\
    \bottomrule
    \end{tabular}
\end{table}

\subsection{Test Cases for Automated Evaluation}

We use automated testing to check correctness of a generated solution. An alternative approach would be to use a reference solution together with either the BLEU~\cite{papineni-etal-2002-bleu} or the ROUGE~\cite{lin-2004-rouge} evaluation scores. However, having in mind how many different solutions may be functionally equivalent, we decided against this approach. Another benefit of test-driven evaluation is that it is easier to create instances for the dataset. This approach is also better to test edge cases for a specific problem, because we can specify multiple test cases for each instance. Finally, as raised by \citet{Chen2021}, functional correctness evaluation is closer to how human developers verify code correctness. 

For each example, we provide between 1 and 4 test cases, including corner cases. All of the test cases contain valid inputs. We do not check incorrect inputs in this version of our benchmark. Provided test cases are meant to check correct execution of the main branch in function logic. We made this decision based on our history of working with AI coding assistants. From our experience, they are more likely to generate useful code if they only generate main operations in the first go. Handling of incorrect input can be then added in next iterations. This, however, is a code infilling task, which we do not cover in this version of our benchmark.

\begin{lstfloat}[b]
\begin{lstlisting}[language=Python, caption=Example of augmentation used in samples creation., label=lst:sample-augmentation]
# operation, lat/lon, gdf
def check_point(lat: float, lon: float, polygons: gpd.GeoDataFrame) -> bool:
    """Check if a point is within any of the polygons. Polygons are stored in a GeoDataFrame."""
---
# operation, lat/lon, shp
def check_point(lat: float, lon: float, polygons: str) -> bool:
    """Check if a point is within any of the polygons. Polygons are stored in the shapefile at given path."""
---
# operation, shapely, gdf
def check_point(point: shapely.geometry.Point, polygons: gpd.GeoDataFrame) -> bool:
    """Check if a point is within any of the polygons. Polygons are stored in a GeoDataFrame."""
---
# operation, shapely, shp
def check_point(point: shapely.geometry.Point, polygons: str) -> bool:
    """Check if a point is within any of the polygons. Polygons are stored in the shapefile at given path."""
---
# semantic, lat/lon, gdf
def check_location(lat: float, lon: float, cities: gpd.GeoDataFrame) -> bool:
    """Check if a given location is in any of the cities. Cities are stored in a GeoDataFrame."""
---
# semantic, lat/lon, shp
def check_point(lat: float, lon: float, cities: str) -> bool:
    """Check if a given location is in any of the cities. Cities are stored in the shapefile at given path."""
---
# semantic, shapely, gdf
def check_point(location: shapely.geometry.Point, cities: gpd.GeoDataFrame) -> bool:
    """Check if a given location is in any of the cities. Cities are stored in a GeoDataFrame."""
---
# semantic, shapely, shp
def check_point(location: shapely.geometry.Point, cities: str) -> bool:
    """Check if a given location is in any of the cities. Cities are stored in the shapefile at given path."""
\end{lstlisting}
\end{lstfloat}

\subsection{Dataset Summary}

In total, our dataset consists of \textbf{20} unique geospatial tasks of varius types. After augmentations within the dimensions described above, we obtained \textbf{77} samples to form our dataset. In the first dimension of categorisation, we have \textbf{13} single-step and \textbf{7} multi-step tasks. Separately, we can summarise the tasks according to the other dimensions - used input, tools, and framing. This is presented in Table~\ref{tab:sample-categories}. Every sample can fit into multiple categories, because it may use more than one tool. Examples of the considered tasks are presented in Listings~\ref{lst:sample-simple}~and~\ref{lst:sample-complex}.

\begin{lstfloat}
\begin{lstlisting}[language=Python, caption=Single operation task with Shapely and GeoJSON., label=lst:sample-simple]
def check_country(
    point: shapely.geometry.Point, 
    countries: str,
) -> str:
    """Check in which country a point is. The countries is the path to the geojson with countries boundaries with `name` feature."""
\end{lstlisting}
\end{lstfloat}

\begin{lstfloat}
\begin{lstlisting}[language=Python, caption=Multi-step task using GeoPandas., label=lst:sample-complex]
import geopandas as gpd

def building_to_parcel_ratio(
    buildings_gdf: gpd.GeoDataFrame, 
    parcels_gdf: gpd.GeoDataFrame,
) -> float:
    """Calculate the mean ratio of the buliding area compared to the parcel it stands on."""
\end{lstlisting}
\end{lstfloat}

\begin{table*}[ht]
    \centering
    \caption{Results with greedy decoding , also including HumanEval scores for reference.}
    \label{tab:res-full}
    \begin{tabular}{lrrrr}
    \toprule
    {} & HumanEval & Accuracy & Pass@1 & Pass\_any@1 \\
    \midrule
    bigcode/starcoder2-7b             & 34.09\%$^1$ & 38.92\% & 32.47\% & 37.66\% \\
    meta-llama/CodeLlama-7b-hf        & 29.98\%$^1$ & 28.74\% & 22.08\% & 31.17\% \\
    meta-llama/CodeLlama-7b-Python-hf & 40.48\%$^1$ & 30.54\% & 24.68\% & 31.17\% \\
    meta-llama/Meta-Llama-3-8B        & 33.50\%$^2$ & 25.75\% & 15.58\% & 28.57\% \\
    mistralai/Mistral-7B-v0.1         & 28.70\%$^2$ & 19.16\% & 14.29\% & 20.78\% \\
    google/gemma-7b                   & 35.40\%$^2$ & 11.98\% & 9.09\% & 11.69\% \\
    google/codegemma-7b               & 40.13\%$^1$ & 17.37\% & 12.99\% & 15.58\% \\
    \bottomrule
    \end{tabular}
    \caption*{\tiny{
    $^1$ According to Big Code Leaderboard: \url{https://huggingface.co/spaces/bigcode/bigcode-models-leaderboard} (7.06.2024)\\
    $^2$ According to EvalPlus Leaderboard: \url{https://evalplus.github.io/leaderboard.html} (7.06.2024),
    }
    }
\end{table*}

\section{An Evaluation of Existing Code LLMs in Geospatial Tasks}

This section describes an evaluation of generic code generation models on our newly created dataset covering geospatial coding problems. The section first describes the experimental setup, where we highlight the models selection and evaluation metrics. Then we present the results of our experiments.

\subsection{Experimental Setup}

\paragraph{Models}

We decided to use a combination of code specific models and base LLMs. All of the models are available on HuggingFace~\cite{Wolf_Transformers_State-of-the-Art_Natural_2020} We used the \textit{transformers} library to load the models and the \textit{bitsandbytes} library\footnote{https://github.com/TimDettmers/bitsandbytes} to run them efficiently, considering 4-bits per parameter. We selected 7 models for our experiments:

\begin{itemize}
    \item \textit{bigcode/starcoder2-7b} \cite{Lozhkov2024};
    \item \textit{meta-llama/CodeLlama-7b-hf} \cite{Roziere2023};
    \item \textit{meta-llama/CodeLlama-7b-Python-hf} \cite{Roziere2023};
    \item \textit{meta-llama/Meta-Llama-3-8B} \cite{llama3modelcard};
    \item \textit{mistralai/Mistral-7B-v0.1} \cite{Jiang2023};
    \item \textit{google/gemma-7b} \cite{Gemma};
    \item \textit{google/codegemma-7b} \cite{CodeGemma}.
\end{itemize}

We argue that 7B/8B parameter versions of the models are worth testing in this scenario, because they can be efficiently used on personal machines. This makes them good candidates as the foundation to fine-tuned versions designed for geospatial tasks, which could be used as AI coding assistants. 

\paragraph{Evaluation metrics}
When testing the correctness of generated code, we used three metrics:

\begin{itemize}
    \item Accuracy - A simple percentage of passed test cases, summed across all samples in our dataset.
    \item Pass@1 - A percentage of completely correct solutions (i.e., passing all test cases) among all samples in our dataset.
    \item Pass\_any@1 - A percentage of solutions which pass at least one test case, again over all dataset instances.
\end{itemize}

We also report pass@1 on the HumanEval dataset according to the Big Code Models Leaderboard\footnote{\url{https://huggingface.co/spaces/bigcode/bigcode-models-leaderboard} (7.06.2024)} (for code models) and the EvalPlus Leadeboard\footnote{\url{https://evalplus.github.io/leaderboard.html} (7.06.2024)} (for generic LLMs). This gives us a reference on how those models compare on generic programming tasks.

\paragraph{Hyperparameters}
Since we evaluate our models via the \textit{pass@1} metric, we used greedy decoding to produce a single output. We set the $max\_length=200$, which we verified to be enough to generate a complete function for all of the models and test cases.

\paragraph{Evaluation pipeline}
For testing, we first have to clean the generated code. We trim the response to only contain a single function. We do this by searching for the second occurrence of the "def " string. To avoid errors which may be a result of missing imports, we searched through the generated answers in search for all libraries that the models wanted to use. We then install them in a virtual environment in their most recent versions. Before running tests, we import them all. By doing this, we make the task easier (i.e., the model does not have to generate imports). However, this is consistent with the prompt format which we selected. Since we generate the code after the docstring, and since it is very uncommon to put imports inside functions, it would be highly unlikely that models generate imports there. We aim to test the capabilities to solve geospatial tasks, not the abilities to generate correct imports in unorthodox places.

\paragraph{Computational resources}
Experiments were conducted on two different machines. First, we used a machine with an Nvidia GTX 1080 with 8GB of VRAM. For running models, we used the \textit{transformers} library and we loaded them in 4-bits quantization using \textit{bitsandbytes}. All of the models fit on this GPU, which means that our experiments could be easily recreated on consumer grade hardware. However, to speed up the computation of final experiments, we used a machine with a single Nvidia A100 80GB card, which allowed us to run code generation in parallel. We made sure to keep the same configuration as on the first machine, to ensure reproducibility.

\subsection{Experimental Results}

\paragraph{Combined results} We report combined results over our dataset in Table~\ref{tab:res-full}. This table shows all of the metrics used in our experiments (accuracy, pass@1, pass\_any@1) alongside HumanEval performance. We can see that there are differences regarding the performance on generic programming tasks compared to geospatial tasks. The best results have been achieved by the \textit{bigcode/starcoder2-7b} model, which was only 4th on generic programming. On the other hand, the \textit{gemma} and \textit{codegemma} models, which report high performance on HumanEval, work poorly on code generation for the geospatial domain. Upon examination of generated code, we found that these models tend to hallucinate and start to generate repetitive blocks of code in the middle of a function.

\paragraph{Edge cases in testing} We included two metrics - pass@1 and pass\_any@1 - to see if models can generate code that is at least partially correct. As expected if we expect only at least one test case to pass, we get higher scores across all models. This result indicates that, for some cases, an incorrect piece of code may still result in correct results for some test cases. This shows that our approach to provide multiple test cases for each sample is correct, and we should further extend our test scenarios. 

\begin{table}[h]
    \centering
    \caption{Results for different task complexity (pass@1).}
    \label{tab:res-task}
    \begin{tabular}{lrr}
    \toprule
     & simple & complex \\
    \midrule
    bigcode\_starcoder2-7b & 45.45\% & 15.15\% \\
    mistralai\_Mistral-7B-v0.1 & 20.45\% & 6.06\% \\
    meta-llama\_CodeLlama-7b-hf & 29.55\% & 12.12\% \\
    google\_gemma-7b & 13.64\% & 3.03\% \\
    google\_codegemma-7b & 20.45\% & 3.03\% \\
    meta-llama\_CodeLlama-7b-Python-hf & 38.64\% & 6.06\% \\
    meta-llama\_Meta-Llama-3-8B & 22.73\% & 6.06\% \\
    \bottomrule
    \end{tabular}
\end{table}

\paragraph{Task complexity} Before running the experiments we speculated that multi step (complex) tasks would be more challenging for code generating models. The experimental results backed this assumption. We can observe a significant drop in performance when comparing results based on task complexity. This is consistent across all models and higher if a model records lower performance overall. This means that models that are worse at geospatial code generation find it especially difficult to solve multi-step tasks. We report these results in Table~\ref{tab:res-task}.

\begin{table}[h]
    \centering
    \caption{Results for different task framings (pass@1).}
    \label{tab:res-framing}
    \begin{tabular}{lrr}
    \toprule
     & operation & semantic \\
    \midrule
    bigcode\_starcoder2-7b & 36.36\% & 24.24\% \\
    mistralai\_Mistral-7B-v0.1 & 12.12\% & 15.15\% \\
    meta-llama\_CodeLlama-7b-hf & 18.18\% & 24.24\% \\
    google\_gemma-7b & 9.09\% & 9.09\% \\
    google\_codegemma-7b & 12.12\% & 9.09\% \\
    meta-llama\_CodeLlama-7b-Python-hf & 27.27\% & 21.21\% \\
    meta-llama\_Meta-Llama-3-8B & 15.15\% & 12.12\% \\
    \bottomrule
    \end{tabular}
\end{table}

\paragraph{Task framing} We have also compared the models' performance based on the task framing dimension. In this case, the results are not as straightforward as with task complexity. The difference in performance is smaller, but most models work better when the instruction clearly states which spatial operation are expected to be used. However, two models work better on tasks with semantic framing. For the worse performing \textit{gemma} model, there was no difference between the framing types. Based on these results, we can say that models have a weak understanding of the geospatial aspect of the problems and data, because they do not always choose correct operations, even if they can solve this task when framed more clearly. The full results are presented in Table~\ref{tab:res-framing}.

\begin{table}[h]
    \centering
    \caption{Results for different input formats (pass@1).}
    \label{tab:res-input}
    \begin{tabular}{lrrr}
    \toprule
     & gdf & shp & geojson \\
    \midrule
    bigcode\_starcoder2-7b & 27.27\% & 31.25\% & 22.73\% \\
    mistralai\_Mistral-7B-v0.1 & 13.64\% & 6.25\% & 4.55\% \\
    meta-llama\_CodeLlama-7b-hf & 27.27\% & 18.75\% & 4.55\% \\
    google\_gemma-7b & 9.09\% & 0.00\% & 0.00\% \\
    google\_codegemma-7b & 13.64\% & 0.00\% & 0.00\% \\
    meta-llama\_CodeLlama-7b-Python-hf & 27.27\% & 12.50\% & 13.64\% \\
    meta-llama\_Meta-Llama-3-8B & 18.18\% & 12.50\% & 0.00\% \\
    \bottomrule
\end{tabular}
\end{table}

\begin{table*}[t]
    \centering
    \caption{Results for different tools usage (pass@1).}
    \label{tab:res-tool}
    \begin{tabular}{lrrrr}
    \toprule
     & Shapely & H3 & OSMNX & MovingPandas \\
    \midrule
    bigcode\_starcoder2-7b & 71.43\% & 75.00\% & 16.67\% & 0.00\% \\
    mistralai\_Mistral-7B-v0.1 & 71.43\% & 25.00\% & 0.00\% & 0.00\% \\
    meta-llama\_CodeLlama-7b-hf & 71.43\% & 50.00\% & 0.00\% & 0.00\% \\
    google\_gemma-7b & 71.43\% & 0.00\% & 0.00\% & 0.00\% \\
    google\_codegemma-7b & 71.43\% & 50.00\% & 0.00\% & 0.00\% \\
    meta-llama\_CodeLlama-7b-Python-hf & 85.71\% & 50.00\% & 0.00\% & 0.00\% \\
    meta-llama\_Meta-Llama-3-8B & 57.14\% & 50.00\% & 0.00\% & 0.00\% \\
    \bottomrule
    \end{tabular}
\end{table*}

\paragraph{Input format} Another dimension focused on passing the same data in different formats widely adopted in the geospatial domain. We tested geodataframes, shapefiles and geojson files, as some of the most popular geospatial data types. Surprisingly, the format of the data makes a difference based on our experimental results. We report results in this scenario in Table~\ref{tab:res-input}. We can generally observe that models work better with geodataframes. Here we see clear room for improvement by LLM finetuning, because the models did not try to load those geospatial files using geopandas. We should obtain better code LLMs with this type of examples in training data. We intend to explore this in our future research, when we focus on improving LLMs on geospatial code generation.

\begin{table}[h]
    \centering
    \caption{Results for different formats for points (pass@1).}
    \label{tab:res-point}
    \begin{tabular}{lrr}
    \toprule
     & lat\_lon & shapely \\
    \midrule
    bigcode\_starcoder2-7b & 33.33\% & 66.67\% \\
    mistralai\_Mistral-7B-v0.1 & 4.76\% & 6.67\% \\
    meta-llama\_CodeLlama-7b-hf & 4.76\% & 33.33\% \\
    google\_gemma-7b & 0.00\% & 6.67\% \\
    google\_codegemma-7b & 4.76\% & 13.33\% \\
    meta-llama\_CodeLlama-7b-Python-hf & 23.81\% & 26.67\% \\
    meta-llama\_Meta-Llama-3-8B & 9.52\% & 20.00\% \\
    \bottomrule
    \end{tabular}
\end{table}

\paragraph{Handling geometries} We differentiated some examples based on the format of the geometry data in the input - in our case points. We either passed the individual data points as two numbers with latitude and longitude values or as Point objects from the \textit{shapely} library. The results are shown in Table~\ref{tab:res-point}. When comparing model performance based on this dimension, we can see that models perform better when working with \textit{shapely} objects. Here we can also see potential for improvements with finetuning, since lat/lon coordinates can be easily converted into objects representing geometries.

\paragraph{Tool knowledge} We have also prepared a set of examples which would test model capabilities in terms of using existing tools, common in the geospatial domain. We present the results in Table~\ref{tab:res-tool}. Our selection is not comprehensive, but we aimed to cover different types of popular operations. We selected:

\begin{itemize}
    \item the \textit{shapely} library for basic geometry manipulations;
    \item \textit{H3} as one of the most popular libraries for geospatial data preprocessing and aggregation;
    \item \textit{OSMNX}, since the OSM is widely used and correspond to a rich source of geospatial data;
    \item \textit{MovingPandas} as an example of a specialised library for complex geospatial data manipulation. We used functionalities for trajectory loading and basic calculations;
\end{itemize}

\begin{lstfloat}
\begin{lstlisting}[language=Python, caption=Placeholder generated by an LLM., label=lst:placeholder]
import geopandas as gpd

def average_speed(trajectory: gpd.GeoDataFrame) -> float:
    """Calculate the average speed of a trajectory in meters per second. It is saved in a geodataframe as a collection of points and timestamps in the `geometry` and `t` columns. Use the `movingpandas` library."""
    # YOUR CODE HERE
    raise NotImplementedError()
\end{lstlisting}
\end{lstfloat}

We can observe that models work only on the \textit{shapely} and \textit{h3} libraries, failing almost completely on the other two libraries. For some models, we observed that they generated placeholders for functions, in the form presented in Listing~\ref{lst:placeholder}. This would indicate that the model does not know this library, and does not attempt to generate any code. We know that this selection of just 4 libraries is not comprehensive, but we think that it shows basic limitations of current LLMs. An AI coding assistant, which is unable to use popular tools is not very useful. We see a path to mitigate this limitation in the future, by building code LLMs designed specifically for geospatial data and tasks.

\section{Conclusions and Future Work}

In this paper, we proposed a new dataset for LLM evaluation on geospatial code tasks. We took inspiration on existing code generation datasets, in terms of task framing and evaluation. We also selected different viewpoints through which we can look at geospatial code generation. Based on this, we manually created a set of tasks that are differentiated based on those viewpoints/dimensions. We have also tested a selection of existing code generation models on our dataset. We found that model performance varies in all of the dimensions, which shows that they are valid viewpoints for assessing geospatial code generation capabilities. 

\paragraph{Limitations}

Our work is just the first steps towards the construction of a comprehensive geospatial code generation benchmark. It has shown that our direction is correct. Based on the multi-dimension task definition that was considered, we can obtain a difficult benchmark for code generation LLMs. However, the current of version of the benchmark should be expanded to cover more typical tasks and tools from the geospatial domain. Due to computational constraints we limited ourselves to 7B/8B scale LLMs, which we also aim to extend in the future.

\paragraph{Future Work}

Based on the results described in this paper, we can see several directions for future work. Firstly, we will extend the dataset to cover more tasks and tools. We have seen that edge case testing is very important, so we plan to add more test cases for each sample. We may follow the approach presented by \citet{evalplus}, and use larger LLMs for this part. Secondly, we would like to include more models and create a leaderboard, which would be easily extendable when new models are released. This would allow the community to keep track of the current performance of code LLMs on geospatial tasks, and see the areas that are the most problematic. Finally, we plan to use the insights from this comparison to train a code LLM specific to the geospatial domain.


\bibliographystyle{ACM-Reference-Format}
\bibliography{references}


\end{document}